\documentclass{article}
\usepackage{spconf,amsmath,graphicx,hyperref}
\usepackage{amssymb,booktabs}
\hypersetup{hidelinks}
\graphicspath{{figures/}}
\title{LithoWorld: An Endpoint World Model for\\Lithography Process-Window Recovery}
\name{Botong Zhao$^{1,2}$, Fang Yu$^{1,2}$, Xinyuan Chen$^{2}$, Yue Lu$^{1}$\thanks{Corresponding author: Yue Lu.}}
\address{$^{1}$Shanghai Key Laboratory of Multidimensional Information Processing,\\
East China Normal University, Shanghai, China\\
$^{2}$Shanghai Artificial Intelligence Laboratory, Shanghai, China}

\begin{document}
\ninept
\maketitle
\begin{abstract}
Lithography process drift can move a fabricated pattern outside its process window, producing unqualified scanning electron microscopy (SEM) morphology. We present LithoWorld, an endpoint world model for SEM-guided parameter recovery. Interior representative targets jointly determine qualified reference appearance and correction supervision. A causal predictive branch aligns current-image features with endpoint representations, while a conditional flow policy generates a two-dimensional correction as an action chunk. The model learns from 173 groups of measured SEMs organized as focus-exposure maps, preserving unmeasured cells as unknown. On 200 test pairs from the 73-group subset, LithoWorld achieves 88.5\% qualified-region success and 46.0\% interior success, exceeding the strongest comparison method by 2.5 and 6.0 percentage points, respectively. Ablations separate the effects of target selection, endpoint alignment, and output parameterization. Qualitative SEM retrieval illustrates recovery to qualified interiors without requiring exact reference-coordinate matching.
\end{abstract}

\section{Introduction}
\label{sec:intro}
The process window in lithography describes operating conditions that produce acceptable pattern morphology \cite{mack2007lithography}. Process drift can move a pattern outside this region even when its intended layout is unchanged. Scanning electron microscopy (SEM) reveals the resulting structural deviation, but identifying an unqualified image does not directly specify how to recover the process parameters. We study this inverse problem using sparsely measured focus-exposure maps (FEMs): infer a correction from the current SEM and a qualified reference, and return to an observed qualified region.

Two issues connect process-window recovery to world modeling. Multiple parameter cells can produce qualified morphology, so matching one designated coordinate is too restrictive as a success criterion. Yet these cells have unequal local support: a nearby qualified point may sit at the window boundary, whereas an interior point is surrounded by qualified observations. Learning an endpoint representation should therefore account for which qualified appearance supplies supervision. Missing cells also prevent treating an enclosing rectangle or convex hull as a verified process window.

\begin{figure*}[t]
  \centering
  \includegraphics[width=\textwidth]{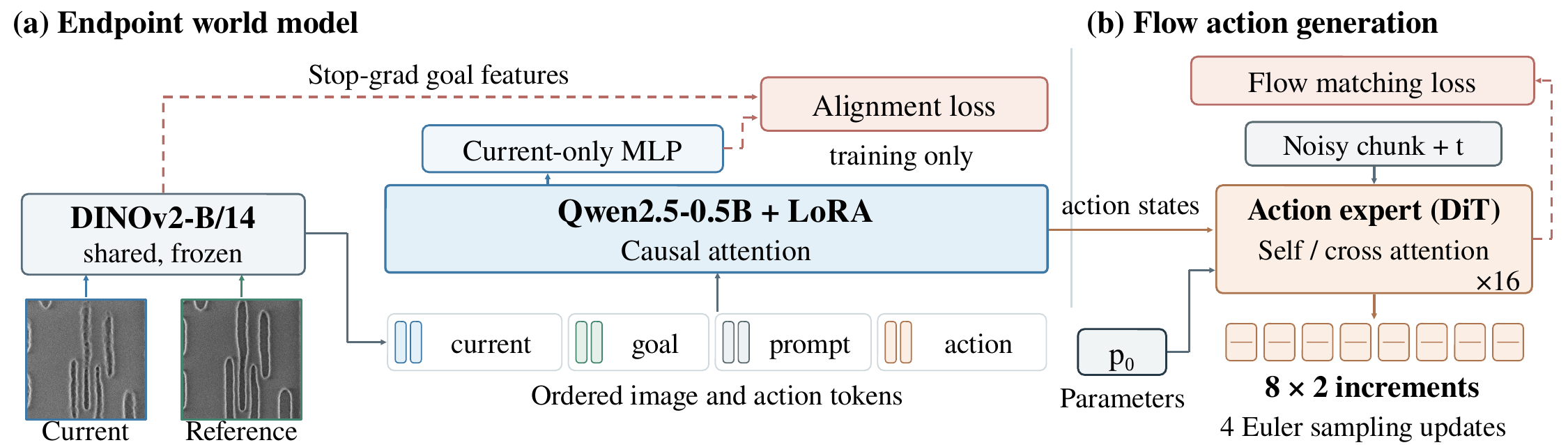}
  \caption{LithoWorld. A shared frozen encoder supplies current/reference SEM tokens to a causal Qwen backbone. Current hidden states predict the qualified endpoint representation through an MLP, with stop-gradient reference features as supervision. Action hidden states, current parameters, and noisy actions condition the DiT expert. Dashed alignment paths are training-only. The reference is selected from an observed process-window interior where available; its coordinates and the full FEM label map are excluded from policy inputs.}
  \label{fig:method}
\end{figure*}

LithoWorld\ couples an \emph{endpoint world model} with a direct flow policy (Fig.~\ref{fig:method}). Here, world modeling means predicting a selected qualified endpoint's visual representation from current-image context. A fully qualified $3\!\times\!3$ neighborhood defines the preferred interior target whenever one exists. That target supplies both the reference SEM and correction label. A causal predictive branch aligns current features with its representation, and a shared policy learns the corresponding correction. Thus, process-window support determines \emph{which outcome to represent}, while endpoint alignment shapes the features used for recovery.

Our contributions are an interior-target supervision rule for sparse lithography FEMs; a causal endpoint world model that regularizes parameter recovery without copying reference tokens; and a grouped evaluation protocol for real SEMs with direct-regression controls. Experiments assess qualified-region recovery and interior placement separately, and SEM lookup cases show how endpoint predictions relate to observed morphology. The study focuses on offline parameter planning from sparse measurements.

\section{Related Work}
\label{sec:related}
\textbf{World models and visual representations.}
DINOv2 provides self-supervised patch features \cite{oquab2024dinov2}, and joint-embedding prediction learns representations by predicting target embeddings \cite{ijepa}. JEPA-WAM couples world modeling with action generation \cite{lin2026jepawam}. We reuse its public action-head infrastructure for static SEM pairs. Our endpoint world model predicts qualified appearance as an auxiliary task; it does not model action-conditioned intermediate fabrication states.

\textbf{Conditional action generation.}
ACT learns action chunks \cite{zhao2023act}; Diffusion Policy uses conditional action diffusion \cite{chi2023diffusionpolicy}. MDT supports image goals \cite{reuss2024mdt}, MoDE uses expert denoisers \cite{reuss2025mode}, and ManiFlow uses consistency flow training \cite{yan2025maniflow}. Our head follows flow matching \cite{lipman2023flow} and the GR00T architecture \cite{nvidia2025gr00t}. We adapt these policy families to SEM conditioning and two-dimensional process correction. Direct two-dimensional regression is essential because the chunk labels derive from one endpoint.

\section{Method}
\label{sec:method}
\subsection{Offline parameter recovery}
Let $I_c$ be the current SEM, $I_g$ a qualified reference from the same map, and $p_0\in[-1,1]^2$ the current parameters normalized using known grid bounds. Given $(I_c,I_g,p_0)$ and a fixed prompt, the policy outputs $A=(a_0,\ldots,a_{H-1})\in\mathbb{R}^{H\times2}$ with $H{=}8$. Its endpoint is
\begin{equation}
 \hat p_H=p_0+\sum_{k=0}^{H-1}a_k.
 \label{eq:endpoint}
\end{equation}
The policy receives neither the target coordinates nor the map's full quality labels. The benchmark assumes access to a qualified reference image; the evaluator uses map annotations offline to construct that input under a common rule for every method.

In the equations, $\mathcal{G}$, $\mathcal{C}$, and $p^\star$ use the same normalized coordinates as $p_0$. Neighborhoods are indexed on the original grid, and evaluation applies the inverse normalization before cell lookup and margin calculation. Let $\mathcal{G}$ contain the centers with observed qualified SEMs. We associate each center with its bounded nearest-center cell, using at most half a grid spacing along each axis. If $U(\mathcal{G})$ denotes the union of qualified cells, success is $\mathbf{1}[\hat p_H\in U(\mathcal{G})]$. Unknown cells do not count as verified success, and out-of-range predictions remain failures without clipping. This definition separates task success from exact recovery of a designated reference coordinate.

\subsection{Interior representative supervision}
For the complete $3\!\times\!3$ neighborhood $\mathcal{N}_3(p)$, define
\begin{equation}
 \mathcal{C}=\{p\in\mathcal{G}:\mathcal{N}_3(p)\subseteq\mathcal{G}\}.
 \label{eq:core}
\end{equation}
Neighbors outside the measured grid or with missing labels cannot satisfy this rule. If $\mathcal{C}\ne\varnothing$, we select the nearest member to $p_0$, breaking distance ties by a larger observed window margin and then a deterministic coordinate order. If $\mathcal{C}=\varnothing$, selection over $\mathcal{G}$ prioritizes margin, local qualified density, and adjustment distance, and records that no fully supported interior target exists. The margin is the $L_\infty$ distance to the boundary of the union of known qualified cells, measured in grid units.

The selected point $p^\star$ supplies an existing image $I_g=I(p^\star)$. Interpolation constructs correction labels only; current and reference SEMs are measured. The imitation target follows straight-line interpolation:
\begin{equation}
 \begin{aligned}
 a_k^\star&=\frac{p^\star-p_0}{H},\quad k=0,\ldots,H-2,\\
 a_{H-1}^\star&=p^\star-p_0-\sum_{k=0}^{H-2}a_k^\star.
 \end{aligned}
 \label{eq:label}
\end{equation}
The final residual corrects numerical accumulation. These increments encode an endpoint plan; no intermediate observation or action duration is available. Target construction changes both the reference appearance and action supervision. Its ablation must therefore be interpreted as a change in the complete target-construction policy.

\begin{table*}[t]
\centering
\caption{Comparison on the FEM dataset using a common frozen DINOv2 encoder. Results are mean $\pm$ sample standard deviation over three runs. GSR: qualified-region success; ISR: interior success; Macro: group-macro GSR. Rates are percentages; mean margin is in grid units, with zero assigned to failures.}
\label{tab:main}
\begin{tabular}{lrrrrrr}
\toprule
Method & \multicolumn{3}{c}{SEM-73 (\%)} & \multicolumn{2}{c}{SEM-100 (\%)} & SEM-73 margin \\
\cmidrule(lr){2-4}\cmidrule(lr){5-6}
 & GSR $\uparrow$ & ISR $\uparrow$ & Macro $\uparrow$ & GSR $\uparrow$ & ISR $\uparrow$ & (grid) $\uparrow$ \\
\midrule
DINOv2 + MLP & 80.0$\pm$1.0 & 26.5$\pm$2.0 & 80.1$\pm$5.1 & 79.2$\pm$0.5 & 23.1$\pm$0.6 & 0.54$\pm$0.02 \\
ACT-goal & 83.0$\pm$1.5 & 33.5$\pm$1.0 & 86.8$\pm$2.5 & 82.6$\pm$0.7 & 26.9$\pm$0.9 & 0.65$\pm$0.02 \\
Diffusion Policy-goal & 85.0$\pm$1.0 & 38.0$\pm$1.5 & 86.1$\pm$2.3 & 84.7$\pm$0.8 & 29.5$\pm$0.9 & 0.72$\pm$0.02 \\
MDT-image & 84.0$\pm$1.0 & 35.5$\pm$1.0 & 84.1$\pm$1.9 & 83.9$\pm$0.7 & 28.0$\pm$0.9 & 0.69$\pm$0.01 \\
ManiFlow-goal & 86.0$\pm$1.5 & 40.0$\pm$1.0 & 86.6$\pm$3.0 & 86.4$\pm$0.8 & 31.6$\pm$0.6 & 0.78$\pm$0.00 \\
MoDE-goal & 84.5$\pm$1.5 & 38.5$\pm$1.5 & 85.0$\pm$1.7 & 85.2$\pm$0.5 & 30.2$\pm$0.9 & 0.74$\pm$0.01 \\
\midrule
w/o endpoint alignment & 85.5$\pm$1.0 & 41.5$\pm$1.0 & 87.3$\pm$3.1 & 85.0$\pm$0.5 & 30.8$\pm$0.6 & 0.81$\pm$0.02 \\
\textbf{LithoWorld (ours)} & 88.5$\pm$1.0 & 46.0$\pm$1.0 & 90.4$\pm$0.9 & 86.0$\pm$0.8 & 32.5$\pm$0.9 & 0.94$\pm$0.01 \\
\bottomrule
\end{tabular}

\end{table*}

\subsection{Causal endpoint world model}
The frozen DINOv2-B/14 encoder $E$ processes both $224\!\times\!224$ images into $256\!\times\!768$ patch features. A shared MLP projects features to width 896, and role embeddings distinguish current and reference tokens. The sequence
\begin{equation}
 [\mathrm{BOS},\,P(E(I_c)),\,P(E(I_g)),\,\mathrm{prompt},\,Q_a]
 \label{eq:tokens}
\end{equation}
enters a causal Qwen2.5-0.5B backbone \cite{qwen2024qwen25} adapted with LoRA \cite{hu2022lora}; $Q_a$ contains 64 learnable action tokens. The action-token hidden states and $p_0$ condition the flow head.

An MLP $g_\theta$ maps the 256 current-token hidden states $h_c$ to endpoint patch features. Since current tokens precede reference tokens, causal attention prevents $h_c$ from accessing $I_g$. The auxiliary loss is
\begin{equation}
 \mathcal{L}_{\rm align}=\frac1{256}\sum_{j=1}^{256}
 \left[1-\cos\left(g_\theta(h_{c,j}),\operatorname{sg}(E(I_g)_j)\right)\right].
 \label{eq:align}
\end{equation}
This branch receives neither $p_0$ nor actions. It encourages endpoint-related visual features while preventing direct copying of the reference input. Multiple qualified outcomes can nevertheless produce ambiguous targets for visually similar current inputs; alignment is an auxiliary regularizer, not a uniquely identifiable physical transition model.

\subsection{Conditional flow objective and inference}
For target chunk $A^\star$, Gaussian noise $\epsilon$, and time $t\in[0,1]$, set $X_t=(1-t)\epsilon+tA^\star$. The velocity field $v_\theta$ is trained using masked mean squared error:
\begin{align}
 \mathcal{L}_{\rm flow}
 &=\mathbb{E}_{A^\star,\epsilon,t}
 \left[\|v_\theta(X_t,t;h_a,p_0)-(A^\star-\epsilon)\|_{M}^{2}\right],\\
 \mathcal{L}&=\mathcal{L}_{\rm flow}+0.5\mathcal{L}_{\rm align}.
 \label{eq:loss}
\end{align}
Here $M$ is the action-validity mask, and time sampling follows the upstream Beta-based transform restricted to $[0,1]$. Inference starts from Gaussian noise and uses four Euler updates to generate the complete $8\!\times\!2$ chunk. Four numerical integration updates are distinct from the eight planned parameter increments. The alignment head is discarded at inference. Training imitates a chosen representative, although evaluation accepts any verified qualified endpoint; the objective does not assign equal loss to every qualified point.

\section{Experiments}
\label{sec:experiments}
\subsection{FEM dataset and evaluation}
All 173 FEM groups contain acquired SEMs, with no simulated or generated images. We retain two subsets: SEM-73 (73 workbooks, 4,334 measured points) and SEM-100 (100 groups, 9,900 measured points), totaling 14,234 points. Their training/validation/test recovery-pair counts are 413/202/200 and 4,514/628/1,363, respectively; FEM-group splits are 51/11/11 and 70/10/20. Eight SEM-73 test morphology groups contain recovery pairs. Grouping uses image hashes and morphology; wafer/lot metadata are unavailable. All 2,915 missing SEM-73 grid locations remain unknown. Interior targets exist for 754/815 SEM-73 and 5,521/6,505 SEM-100 pairs.

Qualified-region success (GSR) accepts any known qualified endpoint. Interior success (ISR) uses all pairs as its denominator; interior availability is 100\% for SEM-73 test pairs and approximately 81.36\% for SEM-100 pairs. Macro-GSR weights evaluated groups equally. The mean observed window margin is measured in grid units, assigning zero to failures. Unknown and out-of-range endpoints remain unsuccessful without clipping. These are discrete process-window proxies.

\begin{figure*}[t]
 \centering
 \includegraphics[width=\textwidth]{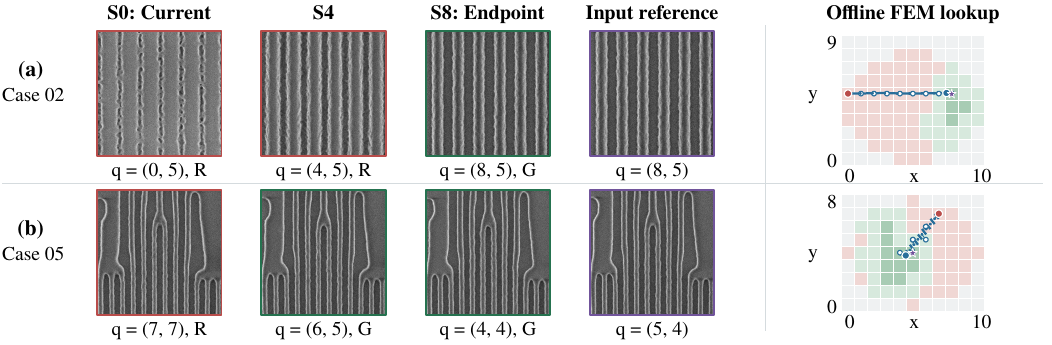}
 \caption{Offline SEM recovery at checkpoint 15,000. Existing SEMs at nearest-grid cells $q$ are shown for S0/S4/S8 of the eight predicted increments. R/G denote source qualification labels. Blue paths show continuous predictions $p$, hollow circles show $q$, and purple stars mark input references; dark green is the observed core and gray is unmeasured. The reference is an input (the same SEM as S8 in (a)). These curated successes use FEM lookup, not newly acquired closed-loop observations; SEM-73 GSR is 89.0\%.}
 \label{fig:cases}
\end{figure*}

\subsection{Training and comparison protocol}
LithoWorld\ is trained on our merged FEM training split using two NVIDIA H200 GPUs. It uses global batch 256 with equal subset sampling, AdamW for 40,000 updates, peak learning rate $2\!\times\!10^{-4}$, 1,200 warmup updates, cosine decay to $10^{-5}$, zero weight decay, and gradient clipping at 1.0. BF16 autocast retains FP32 parameters and optimizer states. LoRA rank/scale/dropout are 32/64/0.1; DINOv2 remains frozen in evaluation mode. The 16-layer flow head has 32 attention heads of width 48; its internal attention width is 1,536, distinct from the configured hidden width of 1,024.

All comparison methods are trained or fine-tuned on the FEM training split and evaluated on the same test split. Table~\ref{tab:main} reports a common frozen-DINOv2 comparison with identical input SEMs, reference selection, parameter normalization, splits, and evaluator. DINOv2+MLP predicts the total correction directly. ACT-goal and Diffusion Policy-goal add reference conditioning. MDT-image uses image goals and disables unavailable language and intermediate-frame objectives; ManiFlow-goal retains consistency flow training; MoDE-goal retains expert routing. All use two-dimensional actions, with horizon eight except the explicit direct-output control. Thus, the names denote adaptations trained for our task.

Tables report the mean and sample standard deviation over three runs. Each method produces one candidate per input, and the evaluator scores its endpoint without test-label reranking. The checkpoint used for qualitative analysis is identified separately so that single-run case statistics are not conflated with the aggregate comparison. The test split was monitored during development; evaluation on independent batches remains an important next step.

\subsection{Comparative results and ablations}
\label{sec:comparisons}

LithoWorld achieves 88.5\% SEM-73 GSR in Table~\ref{tab:main}, compared with 86.0\% for ManiFlow-goal and 80.0\% for direct regression. Its 46.0\% ISR exceeds ManiFlow-goal by 6.0 points, while mean margin rises from 0.78 to 0.94 grid units. Thus, the gain extends beyond reaching a qualified cell to obtaining stronger local window support. Removing endpoint alignment lowers GSR by 3.0 points and ISR by 4.5 points. On SEM-100, LithoWorld attains 86.0\% GSR, slightly below ManiFlow-goal at 86.4\%, but improves ISR from 31.6\% to 32.5\%. The qualification gain is clearest on SEM-73, while interior placement improves on both real-SEM subsets.

\begin{table}[t]
\centering
\caption{Target construction and endpoint alignment on SEM-73. Changing targets changes both reference SEMs and action labels. Rates are percentages; margin is in grid units.}
\label{tab:ablation}
\begin{tabular}{llrrr}
\toprule
Target & Align. & GSR (\%) & ISR (\%) & Margin \\
\midrule
Nearest & Off & 82.5$\pm$1.0 & 13.0$\pm$1.0 & 0.43$\pm$0.02 \\
Nearest & On & 84.5$\pm$1.5 & 16.0$\pm$1.5 & 0.48$\pm$0.01 \\
Interior & Off & 85.5$\pm$1.0 & 41.5$\pm$1.0 & 0.81$\pm$0.02 \\
Interior & On & 88.5$\pm$1.0 & 46.0$\pm$1.0 & 0.94$\pm$0.01 \\
\bottomrule
\end{tabular}

\end{table}

With alignment enabled, interior targets in Table~\ref{tab:ablation} increase GSR from 84.5\% to 88.5\%, ISR from 16.0\% to 46.0\%, and margin from 0.48 to 0.94 grid units. The larger ISR gain matches the target rule's purpose. Alignment adds 2.0 GSR points with nearest targets and 3.0 with interior targets. The two components yield complementary gains: interior supervision selects a more supported destination, while alignment improves recovery under either target rule. The aggregate means alone do not establish statistical interaction.

\begin{table}[t]
\centering
\caption{Output, backbone, and training-data controls evaluated on SEM-73. $H=1$ keeps the same backbone and predicts the total correction. Other rows change one factor.}
\label{tab:design}
\begin{tabular}{lrr}
\toprule
Variant & GSR (\%) & ISR (\%) \\
\midrule
\textbf{LithoWorld (ours)} & 88.5$\pm$1.0 & 46.0$\pm$1.0 \\
$H=1$, same backbone & 88.0$\pm$1.5 & 45.0$\pm$2.0 \\
Small fusion replaces Qwen & 85.5$\pm$1.5 & 41.0$\pm$1.0 \\
SEM-73-only training & 86.5$\pm$1.5 & 43.0$\pm$1.5 \\
SEM-100-only training & 74.5$\pm$1.5 & 30.0$\pm$2.0 \\
Without reference image & 77.0$\pm$1.0 & 31.0$\pm$1.0 \\
\bottomrule
\end{tabular}

\end{table}

The $H=1$ control in Table~\ref{tab:design} reaches 88.0\% GSR, only 0.5 points below $H=8$. This small gap indicates that chunking primarily provides an output parameterization; it does not establish a benefit from temporal feedback. Replacing Qwen with small fusion reduces GSR to 85.5\%, indicating a contribution from the richer cross-image conditioning. Joint training exceeds SEM-73-only training by 2.0 points; SEM-100-only training reaches 74.5\% on SEM-73. These results assess the benefit of additional measured SEMs and transfer between subsets. Equal subset sampling changes SEM-73 exposure, which also affects this comparison. Removing the reference lowers GSR to 77.0\%.

\subsection{Observed results and SEM recovery cases}
The checkpoint-15,000 archive provides SEM-73 GSR of 89.0\% (178/200), ISR of 43.0\% (86/200), macro-GSR of 93.75\%, and mean margin of 0.9244 grid units. SEM-100 yields 86.57\% GSR and 28.32\% ISR. These statistics describe the checkpoint used in Fig.~\ref{fig:cases}; Table~\ref{tab:main} reports the separate three-run aggregate. SEM-73 failures are unevenly distributed: one 48-pair group accounts for 21 of 22 failures. The higher macro score therefore does not imply uniform reliability. Across monitored checkpoints, SEM-73 GSR peaks at 89.5\% at update 22,500, while ISR peaks at 54.5\% at update 7,500; qualification and interior placement favor different checkpoints.

Figure~\ref{fig:cases} shows two morphologically different SEM-73 cases from the fixed 15,000-update archive. Eight predicted increments are accumulated without resampling or replanning. Each displayed SEM is retrieved from the same FEM at $q=\lfloor p+0.5\rfloor$ on each grid axis. The line-pattern case first matches a qualified cell at S7; the branched-pattern case does so at S4. Both final cells are interior, with observed margins of 1.0891 and 1.7190 grid units. In (b), the endpoint cell $(4,4)$ differs from the reference cell $(5,4)$, illustrating process-window recovery without exact target-coordinate matching. They were selected from successful cases and do not establish monotonic defect reduction or population performance.

\subsection{Scope}
The world model regularizes endpoint representations; it lacks action-conditioned intermediate SEM supervision. Grid axes have no physical dose/focus calibration, and lookup images are not measurements newly acquired at the continuous predictions. Deployment also requires a practical source of the qualified reference. If historical qualified coordinates are available, direct reuse of those coordinates is an additional baseline. Independent batches, calibrated process axes, and group-level uncertainty estimates are needed to assess transfer beyond this offline benchmark.

Concentrated SEM-73 failures motivate reporting both pair-weighted and group-weighted scores: strong performance on small groups can obscure failures on a frequent morphology. Macro-GSR should therefore accompany GSR and per-group counts. ISR tests discrete core membership, whereas margin measures clearance within the qualified-cell union. These complementary metrics distinguish qualification, interior placement, and morphological variation, but neither certifies tolerance to unmeasured physical perturbations.

\section{Conclusion}
LithoWorld\ connects endpoint world modeling with lithography process-window recovery through shared interior targets. Its direct policy learns SEM-conditioned corrections while evaluation accepts any observed qualified endpoint. On the SEM-73 test pairs, it achieves 88.5\% qualified-region success and 46.0\% interior success. The ablations show complementary benefits from interior target selection and endpoint alignment, while the direct-output control clarifies the role of action chunking. Future work will examine calibrated process coordinates and feedback from newly acquired SEMs.

\clearpage
\bibliographystyle{IEEEbib}
\bibliography{references}
\end{document}